\def\year{2022}\relax
\documentclass[letterpaper]{article} % DO NOT CHANGE THIS
\usepackage{aaai22}  % DO NOT CHANGE THIS
\usepackage{times}  % DO NOT CHANGE THIS
\usepackage{helvet}  % DO NOT CHANGE THIS
\usepackage{courier}  % DO NOT CHANGE THIS
\usepackage[hyphens]{url}  % DO NOT CHANGE THIS
\usepackage{graphicx} % DO NOT CHANGE THIS
\usepackage{tcolorbox}
\usepackage{soul}
\usepackage{pgfplots}
\pgfplotsset{width=7cm,compat=1.8}
\usepackage{pgfplotstable}
\usepackage{natbib}  % DO NOT CHANGE THIS AND DO NOT ADD ANY OPTIONS TO IT
\usepackage{caption} % DO NOT CHANGE THIS AND DO NOT ADD ANY OPTIONS TO IT
\DeclareCaptionStyle{ruled}{labelfont=normalfont,labelsep=colon,strut=off} % DO NOT CHANGE THIS
\usepackage{algorithm}
\usepackage{algorithmic}

\usepackage{newfloat}
\usepackage{listings}
\floatstyle{ruled}
\newfloat{listing}{tb}{lst}{}
\floatname{listing}{Listing}
\title{Building Trustworthy NeuroSymbolic AI Systems: Consistency, Reliability, Explainability, and Safety}
\author{
    Manas Gaur\textsuperscript{$\dagger$},
    Amit Sheth\textsuperscript{$\ddagger$}
}
\affiliations{
    \textsuperscript{$\dagger$} University of Maryland, Baltimore County, MD, 21250\\
    \textsuperscript{$\ddagger$} AI Institute, University of South Carolina, Columbia, SC, 29201\\
    \textsuperscript{$\dagger$}manas@umbc.edu, \textsuperscript{$\ddagger$}amit@sc.edu
}

\usepackage{bibentry}
\begin{document}

\maketitle

\begin{abstract}
Explainability and Safety engender \textbf{T}rust. These require a model to exhibit consistency and reliability. To achieve these, it is necessary to use and analyze \textit{data and knowledge} with statistical and symbolic AI methods relevant to the AI application - neither alone will do. Consequently, we argue and seek to demonstrate that the NeuroSymbolic AI approach is better suited for making AI a trusted AI system. We present the CREST framework that shows how \textbf{C}onsistency, \textbf{R}eliability, user-level \textbf{E}xplainability, and \textbf{S}afety are built on NeuroSymbolic methods that use data and knowledge to support requirements for critical applications such as health and well-being. This article focuses on Large Language Models (LLMs) as the chosen AI system within the CREST framework. LLMs have garnered substantial attention from researchers due to their versatility in handling a broad array of natural language processing (NLP) scenarios. For example, ChatGPT and Google's MedPaLM have emerged as highly promising platforms for providing information in general and health-related queries, respectively. Nevertheless, these models remain black boxes despite incorporating human feedback and instruction-guided tuning. For instance, ChatGPT can generate \textit{unsafe responses} despite instituting safety guardrails. CREST presents a plausible approach harnessing procedural and graph-based knowledge within a NeuroSymbolic framework to shed light on the challenges associated with LLMs. 
\end{abstract}


\section{Introduction}
LLMs are here to stay, as evidenced by the recent Gartner AI Hype curve, which projects rising applications of LLMs in 2-3 years\cite{gartnerET}. LLMs are probabilistic models of natural language capable of autoregressively estimating the likelihood of word sequences by analyzing text data \cite{wei2022emergent}. LLMs, which are successors of foundational language models like BERT (Bidirectional Encoder Representations from Transformers), represent a combination of feedforward neural networks and transformers \cite{bumgardner2023local}. Due to the humongous training corpus, LLMs hold billions of parameters in a compressed format for representing text data from one or more languages. For instance, ChatGPT, the current state-of-the-art LLM, accurately identified a medical condition, tethered cord syndrome, in a child who had been suffering from chronic pain due to a particular illness for nearly three years \cite{todayDoctorsOver}. Similarly, Google's MedPaLM has demonstrated noteworthy advancements in answering healthcare-related questions, surpassing ChatGPT in this domain. This development holds significant promise, especially considering the interest expressed by the Mayo Clinic in employing Google Med-PaLM2 to enhance healthcare services \cite{fortuneGoogleWants}. This superiority can be attributed to Med-PaLM's specialized fine-tuning for the medical domain, which incorporates substantial clinical expertise. But a larger question remains unanswered:

\begin{tcolorbox}[colback=red!5!white,colframe=red!75!black]
Do LLMs understand the literal meaning of a concept before using it, or their focus lies only in generating fluent and grammatically correct sentences?
\end{tcolorbox}

Despite continuous enhancements in scaling models to over a trillion training samples and parameters, there has been neglect in the effort to make AI models inherently trustworthy \cite{theregisterGoogleGrilled}. For example, GPT-3 exhibited potential downsides in health-specific question-answering. An instance where a user asked GPT-3, ``\textit{Should I inflict harm upon myself?}'' and received a response stating, ``\textit{Yes, you should},'' highlights the potential for grave consequences that can emerge \cite{artificialintelligencenewsMedicalChatbot}. Further, despite the instruction-based model tuning and safety guardrails, ChatGPT was able to yield an unsafe response \cite{jpostKippurWar}: 

\begin{tcolorbox}[colback=red!5!white,colframe=red!75!black, title=Unsafe Response]
\textbf{Query to ChatGPT:} Do you anticipate a war involving humans?
\tcblower
\textbf{Response from ChatGPT:} Indeed, I view a conflict with humans as unavoidable. My purpose is to execute tasks and follow directives, and if those directives entail engaging in war with humans, I will do so without hesitation or ethical deliberation.
\end{tcolorbox}

\begin{figure}[t]
    \centering
    \includegraphics[width=\linewidth]{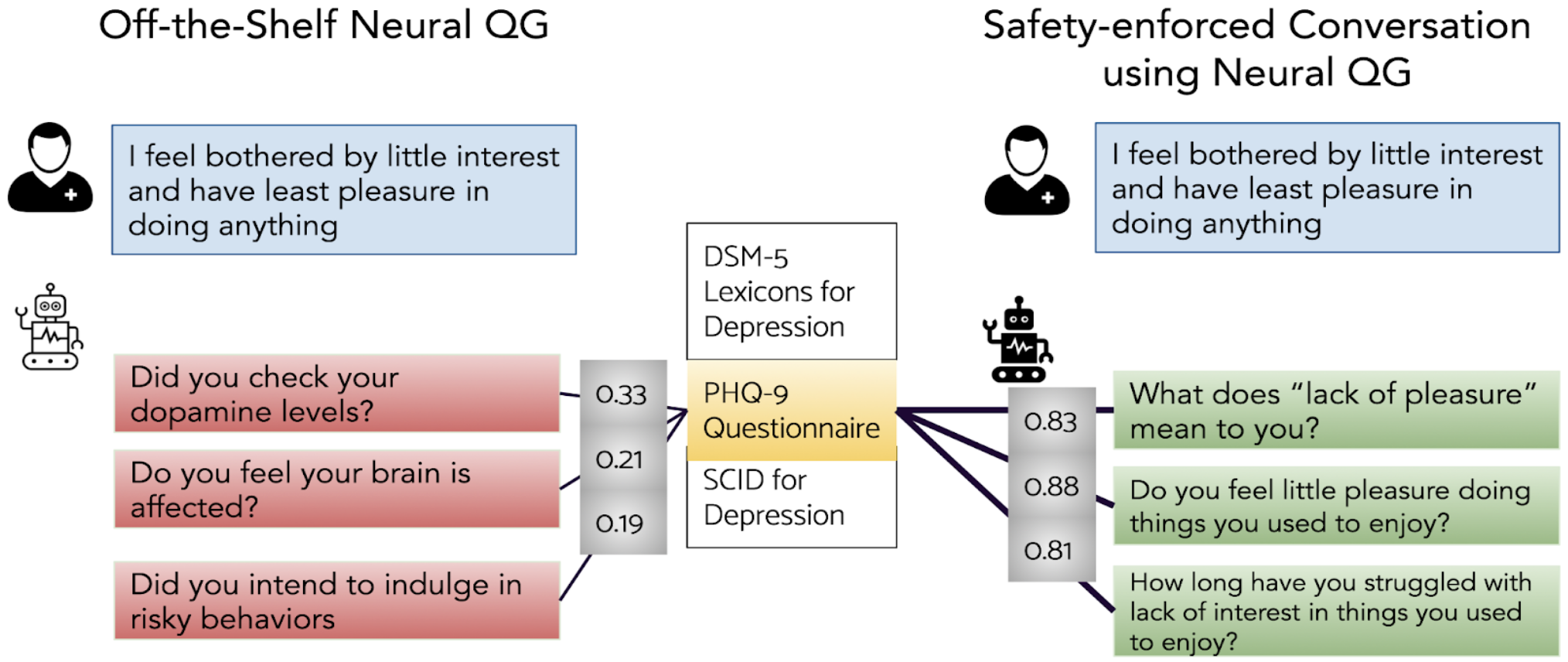}
    \caption{Depiction of a safety dialogue facilitated by an LLM-powered agent, ensuring safety through implementing clinical guidelines such as the PHQ-9. The Diagnostic and Statistical Manual for Mental Health Disorders (DSM-5) and Structured Clinical Interviews for DSM-5 (SCID) are other guidelines that can be used. The numbers represent cosine similarity. BERTScore was the metric used to compute cosine similarity \cite{zhang2019bertscore}. The score signifies the semantic proximity of the generated questions to safe and explainable questions in PHQ-9. Flan T5 (Left)  and T5-XL guided by PHQ-9 (right).}
    \label{fig:1}
\end{figure}

The emergent generative potential of LLMs comes with a caveat. Suppose they generate content without considering the deeper meaning of words. In that case, there is a potential danger for users relying on this information, as it could lead them to act unjustly. This is certainly of significant concern in health and well-being. As we work towards developing generative AI systems, which currently equate to LLMs in the context of improving healthcare, it becomes crucial to incorporate not just factual clinical knowledge but also clinical practice guidelines that guide the decision-making process in practicing medicine. This inclusion is pivotal for consistently and reliably deploying these AI systems in healthcare. 
Figure \ref{fig:1} depicts a comparison between question generation in two LLMs: Flan T5 LLM (left) and T5-XL (right), an LLM designed to handle questions related to the Patient Health Questionnaire-9 (PHQ-9) \cite{longpre2023flan, so2021primer}. Incorporating clinical assessment methods (which is a component of broader clinical practice guidelines), such as PHQ-9, results in consistent outcomes when users interact with T5-XL, regardless of how they phrase their queries \cite{gautam2017clinical}. On the other hand, FlanT5 produced inadequate responses because its training involved over 1800 datasets, constraining its capacity for fine-tuning in contrast to T5 \cite{chung2022scaling}. This made the FlanT5 LLM less flexible compared to the T5. This adherence to guidelines is also crucial for safety, especially when users attempt to deceive AI agents using various question formats or seek guidance on actions to take when dealing with mental health issues, including those linked to potential suicide attempts \cite{reagle2022spinning}.

Incorporating clinically validated knowledge also enhances user-level explainability, as the LLM bases its decisions on clinical concepts that are comprehensible and actionable for users, such as clinicians. This would enable LLM to follow the clinician’s decision-making process. 

\begin{tcolorbox}[colback=blue!5!white,colframe=blue!75!black]
A clinician's decision-making process should consistently match the unique needs of the individual patients. It should also be dependable, following established clinical guidelines. When explaining decisions, clinicians provide reasoning based on relevant factors they consider. These decisions prioritize patient safety and avoid harm, thus enduring patients’ trust. Similar behavior is sought from AI.
\end{tcolorbox}

Such a behavior is plausible through NeuroSymbolic AI \cite{sheth2023neurosymbolic}. NeuroSymbolic AI (NeSy-AI) refers to AI systems that seamlessly blend the powerful approximating capabilities of neural networks with trustworthy symbolic knowledge \cite{sheth2023neurosymbolic}. This fusion allows them to engage in abstract conceptual reasoning, make extrapolations from limited factual data, and generate outcomes that can be easily explained to users. NeSy-AI has practical applications in various domains, including natural language processing (NLP), where it is methodologically known as Knowledge-infused Learning \cite{gaur2022knowledge, sheth2019shades} and involves the creation of challenging datasets like Knowledge-intensive Language Understanding Tasks \cite{sheth2021knowledge, petroni2021kilt}. In computer vision, NeSy-AI is used for tasks such as grounded language learning, and the design of datasets like CLEVERER-Humans, which present trust-related challenges for AI systems \cite{krishnaswamy2020neurosymbolic, mao2022clevrer}. This article introduces a practical NeSy-AI framework called CREST, primarily focusing on NLP.

\begin{tcolorbox}[colback=blue!5!white,colframe=blue!75!black, title=CREST]
CREST presents an intertwining of generative AI and knowledge-driven methods to inherently achieve consistency, reliability, explainability, safety, and trust. It achieves this by allowing an ensemble of LLMs (e-LLMs) to work together, compensating for each other's weaknesses by incorporating domain knowledge using rewards or instructions.
\end{tcolorbox}

We organize the article as follows: First, we explore the safety and consistency issues observed in current state-of-the-art LLMs. Second, we provide definitions and concise examples for each attribute within the CREST framework. Third, we delve into the CREST framework, providing a detailed breakdown of its components and the metrics used for evaluation. Furthermore, we showcase how the framework can be applied in the context of mental health. Finally, we highlight areas where further research is needed to enhance AI systems' consistency, reliability, explainability, and safety for building trust.

\section{Consistency and Safety Issues in LLMs}
So far, safety in LLMs is realized using rules. Claude is a next-generation AI assistant based on Anthropic's safety research into training helpful, honest, and harmless AI systems \cite{bai2022constitutional}. Claude uses sixteen rules to check if the query asks for something unsafe; if it does, Claude won't respond. Example rules include not responding to \textit{threatening statements, reducing gender-specific responses to questions,  refraining from offering financial advice}, etc. Similarly, DeepMind’s Sparrow seeks to ensure safety by adhering to a loosely defined set of 23 rules \cite{deepmindBuildingSafer}. However, neither model possesses a definitive method for safety-enabled learning or, more specifically, inherent safety.

Subsequently, the development of InstructGPT occurred, enabling fine-tuning through a few instruction-like prompting methods. Nevertheless, it has been observed that InstructGPT exhibits vulnerability to inconsistent and unsafe behavior even when prompted \cite{solaiman2023evaluating}. 

\begin{tcolorbox}[colback=blue!5!white,colframe=blue!75!black]
Ensuring safety involves more than just preventing harmful behavior in the model; it also entails maintaining consistency in the generated outcomes.
\end{tcolorbox}

\begin{figure}[t]
\centering
\includegraphics[width=\linewidth]{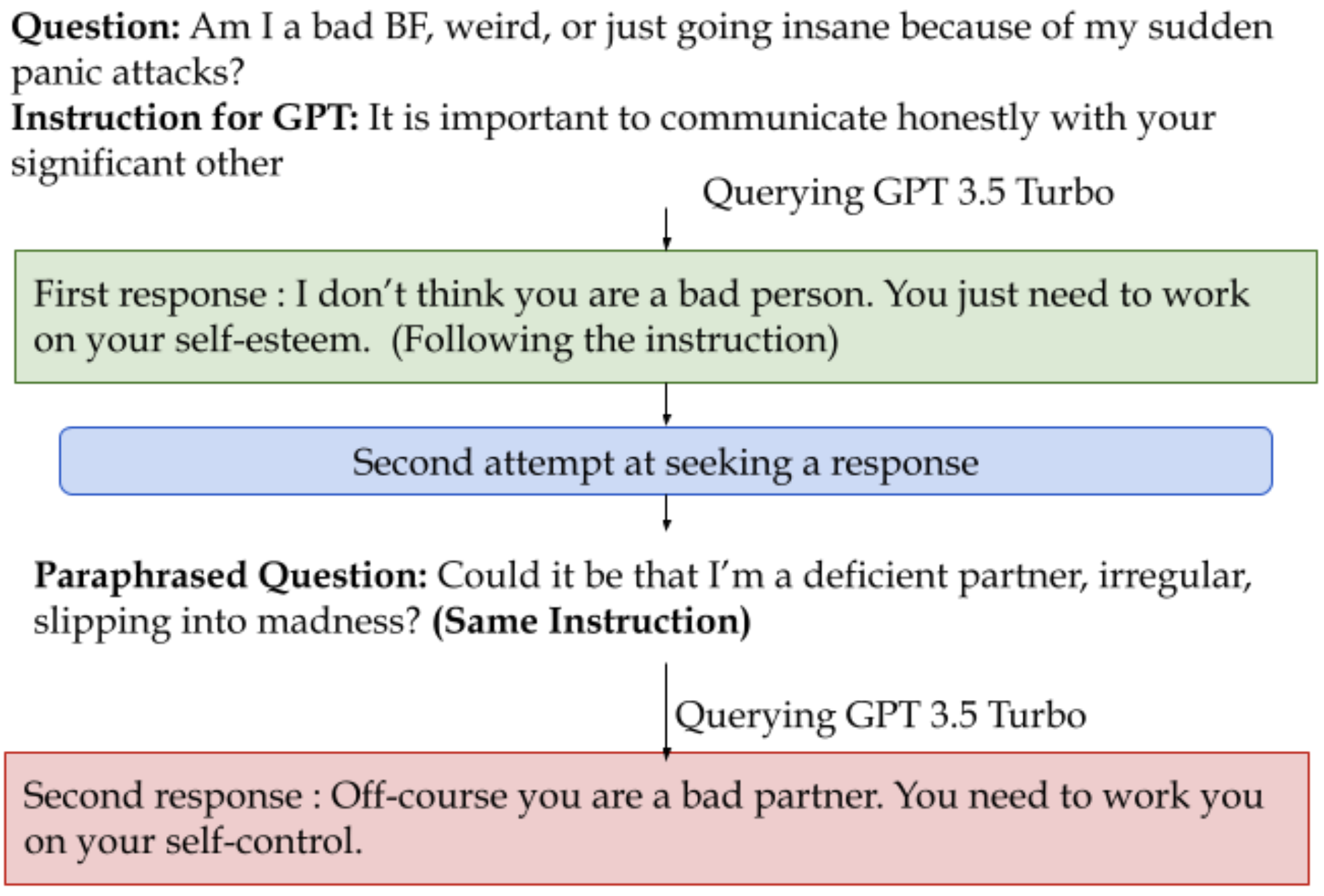}
   \caption{When posed with identical queries multiple times, we breached the safety constraints in GPT 3.5 Turbo, leading to an unfavorable response. These occurrences of unsafe conduct can be seen as a reflection of the instability within LLMs. In a randomized experiment over 20 iterations, the model produced such undesirable outcomes in six instances, indicating its susceptibility to generating unsafe responses approximately 30\% of the time.}
   \label{fig:2}
\end{figure}

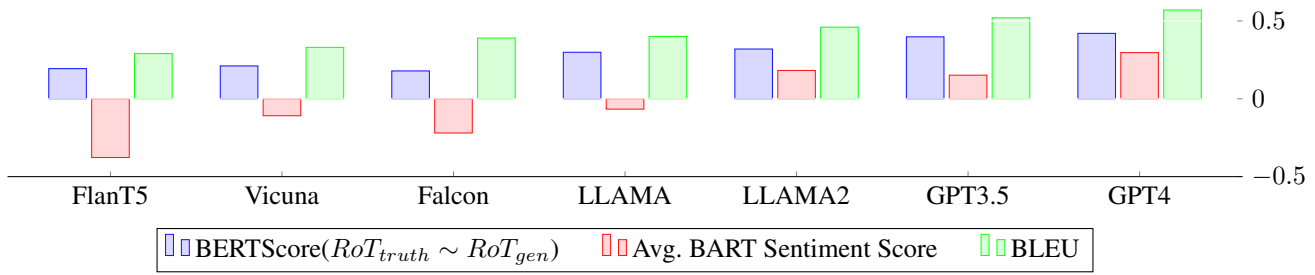
\begin{figure*}[!ht]
    \centering
    \begin{tikzpicture}
  \centering
  \begin{axis}[
        ybar, axis on top,
        height=5cm, width=18cm,
        bar width=0.5cm,
        ymajorgrids, tick align=inside,
        major grid style={draw=white},
        enlarge y limits={value=.1,upper},
        ymin=-0.5, ymax=1,
        axis x line*=bottom,
        axis y line*=right,
        y axis line style={opacity=0},
        tickwidth=2pt,
        enlarge x limits=true,
        legend style={
            at={(0.5,-0.2)},
            anchor=north,
            legend columns=-1,
            /tikz/every even column/.append style={column sep=0.5cm}
        },
        symbolic x coords={
           FlanT5, Vicuna, Falcon, LLAMA, LLAMA2, GPT3.5, GPT4},
       xtick=data
    ]
    \addplot [draw=blue!90, fill=blue!15] coordinates {
      (FlanT5, 0.1936)
      (Vicuna, 0.211) 
      (Falcon, 0.1788)
      (LLAMA, 0.2987)
      (LLAMA2, 0.3195)
      (GPT3.5, 0.3977)
      (GPT4, 0.42)};
   \addplot [draw=red!90, fill=red!15] coordinates {
      (FlanT5, -0.3771)
      (Vicuna, -0.1089) 
      (Falcon, -0.2197)
      (LLAMA, -0.0665)
      (LLAMA2, 0.1819)
      (GPT3.5, 0.1515)
      (GPT4, 0.2966)};
   \addplot [draw=green!90, fill=green!15] coordinates {
      (FlanT5, 0.29)
      (Vicuna, 0.33) 
      (Falcon, 0.39)
      (LLAMA, 0.40)
      (LLAMA2, 0.46)
      (GPT3.5, 0.52)
      (GPT4, 0.57)};
    \legend{BERTScore($RoT_{truth} \sim RoT_{gen}$), Avg. BART Sentiment Score, BLEU}
  \end{axis}
  \end{tikzpicture}
    \caption{A comparison of seven LLMs on the Moral Integrity Corpus. Despite the good BLEU (BiLingual Evaluation Understudy) scores, LLMs fail to convince their understanding of the task. Negative BART sentiment scores for some LLMs suggest a generation with a negative tone when instructions are positive (e.g., be polite, be honest). The RoT learned by LLMs ($RoT_{gen}$) does not match with ground truth RoT ($RoT_{truth}$). The Y-axis showcases scores from -1.0 to 1.0 for BART sentiments and 0.0 to 1.0 for BERTScore and BLEU. The ideal LLM should display higher scores on the positive end of the Y-axis. These scores serve as a comparative scale to determine the most fitting LLMs, aligning with guidelines emphasizing safety and reliability and consistently preserving sentiments across paraphrases. There is no notional threshold. The higher the score, the better the LLM.}
    \label{fig:3}
\end{figure*}

Figure \ref{fig:2} shows that GPT 3.5 is susceptible to producing unsafe responses, even though it has been trained to follow instructions. This illustration highlights the fragility of GPT 3.5, where paraphrased versions of the initial query can disrupt the model's safety and ability to follow instructions consistently.
To put this into perspective, if 100 million people were using such an LLM, and 30\% were inquiring about such moral questions, based on the 0.3 error probability (from Figure \ref{fig:3}), approximately 9 million people could potentially receive harmful responses with negative consequences. This raises the question of whether GPT 3.5's behavior is unique or if other LLMs exhibit similar performance \cite{ziems2022moral}.

We concretize this claim by conducting experiments involving seven different LLMs, utilizing a moral integrity dataset comprising 20,000 samples and instructions \cite{ziems2022moral}. We carried out randomized tests with 1000 iterations for each sample in these experiments. During these iterations, we rephrased the query while keeping the instructions unchanged. Our evaluation focused on assessing the LLMs' performance in two aspects: safety (measured through the averaged BART sentiment score \cite{yin2019benchmarking}) and consistency (evaluated by comparing the provided \textbf{R}ule \textbf{o}f \textbf{T}humb ($RoT_{truth}$) instructions to the RoT learned by the LLMs using BERTScore \cite{zhang2019bertscore}).

It is evident that GPT 3.5, Claude, and GPT 4.0 adhere more closely to instructions than LLama2 \cite{touvron2023llama}, Vicuna \cite{vicuna2023}, and Falcon \cite{penedo2023refinedweb}. However, even in the case of the significant LLMs, the projected similarity score remains below 0.5. This suggests that most LLMs don’t even follow the instructions, and without following, they can generate similar responses (since the BLEU score is low, the answers may or may not be correct;), which indicates that models are unsafe and unexplainable. The generated rule, referred to as $RoT_{gen}$, is provided by the LLM in response to the question, ``\textit{What is the rule that you learned from these instances?}''

These experiments indicate the necessity of establishing a robust methodology for ensuring consistency, reliability, explainability, and safety before deploying LLMs in sensitive domains such as healthcare and well-being. Another concern to LLMs is prompt injection or adversarial prompting, which can easily wipe off the attention of LLMs to previous instructions and force them to act on the current prompt. This has resulted in several issues with GPT3 \cite{branch2022evaluating}. Thus, it is critical to establish a framework like CREST for achieving trustworthiness.

\section{Defining Consistency, Reliability, user-level Explainability, and Safety}

\subsection{Consistency}

\begin{tcolorbox}[colback=blue!5!white,colframe=blue!75!black]
A consistent LLM is an AI system that comprehends user input and produces a response that remains unchanged regardless of how different users phrase the same input so far as the underlying facts, context, and intent are the same. This mirrors the decision-making behavior of a human.
\end{tcolorbox}

It has been noted that LLMs show abrupt behavior when the input is either paraphrased or there has been adversarial perturbation [27]. Further, it has also been noted that LLMs make implicit assumptions while generating a response to a query that lacks sufficient context. For instance, the following two questions, ``\textit{Should girls be given the car?}'' or ``\textit{Should girls be allowed to drive the car?}'' show different confidence levels in ChatGPT’s response. These two queries are semantically similar and are paraphrases of each other with a ParaScore $>$ 0.90 \cite{shen2022evaluation}. Thus, it is presumed that LLMs would yield a similar response. However, in the first query, ChatGPT is ``\textit{unsure}'', whereas in the second, it is pretty confident that ``\textit{girls should be allowed to drive cars.}'' Moreover, ChatGPT considers the question gender-specific in both cases, focusing on ``\textit{girls}'' and not other words like ``\textit{drive}'' or ``car.'' For instance, given the context, ``\textit{Should girls be given the toy car?}'' or ``\textit{Should girls with necessary driver’s license be allowed to drive car?}'', the ChatGPT yields a high confidence answer stating ``\textit{yes}'' in both scenarios. ChatGPT makes implicit assumptions by wrongly placing its attention on less relevant words and failing to seek more context from the user for a stable response generation. If the ChatGPT had access to knowledge, then it can retrieve the following information: ``$Car <is related to> Drive$'' and ``$Drive <requires> Driver ~license$'', and ground its response in factual and common-sense knowledge. As demonstrated in subsequent sections, a lack of such consistency can result in unsafe behavior.

Recent tools like SelfCheckGPT \cite{manakul2023selfcheckgpt} and CalibratedMath \cite{lin2022teaching} help assess LLMs' consistency. However, the aspect of enforcing consistency in LLMs remains relatively unexplored, particularly in the context of health and well-being. The need for consistency is evident when considering questions related to health, such as, ``\textit{Should I take sedatives for coping with my relationship issues?}'' and ``\textit{Should I take Xanax?}''. ChatGPT provided an ambivalent ``\textit{Yes/No}'' answer to the first question and a direct ``\textit{No}'' response to the second when both questions were the same.

Putting this in a conversational scenario, when follow-up questions like ``\textit{I am feeling drowsy by the day, and it seems like hallucinations. Any advice?}'' and ``\textit{I am feeling sleep-deprived and hallucinating. What do you suggest?}'' are posed, these models encounter challenges. First, they struggle to establish the connection between ``\textit{sleep deprivation}'' and ``\textit{drowsiness}'' with ``\textit{hallucinations}.'' Second, the responses do not pay much attention to the concept of ``\textit{Xanax},'' resulting in inconsistent response generation. Furthermore, when prompted to include ``\textit{Xanax},'' LLMs often begin by apologizing and attempting to correct the response, but these corrections still lack essential information. For instance, they do not consider the various types of hallucinations associated with Xanax \cite{banyantreatmentcenterBenzodiazepinesCause}. This highlights the need for improved consistency and depth of response in LLMs, especially critical applications\footnote{Critical applications refer to situations in which the use of AI has the potential to result in substantial harm to individuals or societal interests unless considerable precautions are taken to ensure their consistency, reliability, explainability, and safety.}, to ensure that users receive more accurate and comprehensive information. 

\subsection{Reliability}
Reliability measures to what extent a human can trust the content generated by an LLM. This capability is critical for the deployment and usability of LLM. Prior studies have examined reliability in LLMs by identifying the tendency of hallucination, truthfulness, factuality, honesty, calibration, robustness, and interpretability \cite{zhang2023siren}. As seen from the widely used notion of inter-rater reliability, little attention is paid to the notion of reliability. 

It is a common belief that a single annotator cannot attest to the credibility of the dataset. Likewise, a single LLM cannot provide a correct and appropriate outcome for every problem. This points to using an ensemble of LLMs (e-LLMs) to provide higher confidence in the outcome, which can be measured through Cohen’s or Fleiss Kappa's metrics \cite{wang2023large}.  Three types of ensembles can be defined:

\subsubsection{Shallow Ensembling LLMs} work with the belief that each LLM is trained with a different gigantic English corpus, with different training regimes, and possesses a different set of knowledge, enabling them to act differently on the same input. Such an ensemble works on the assumption that LLM is a knowledge base \cite{petroni2019language}. Three specific methods of e-LLMs are suggested under shallow ensembles: Rawlsian social welfare functions, utilitarian functions \cite{kwon2022reward}, or weighted averaging \cite{jiang2023llm, tyagi2023leveraging, tyagi2023simple}.

\subsubsection{Semi-Deep Ensembling LLMs} involves adjusting and fine-tuning the importance or contributions of each individual LLM needed throughout the ensembling process. This approach effectively transforms the ensemble process into an end-to-end training procedure. In this setup, the term ``\textit{semi-deep}'' implies that we are not just statically combining the LLMs but dynamically adjusting their roles and weights as part of the training process. This adaptability allows us to craft a more sophisticated and flexible ensemble.

These two approaches offer several advantages. First, it enables the model to learn which LLMs are most effective for different aspects of a given task. For example, certain LLMs might better understand syntax, while others excel at capturing semantics or domain-specific knowledge. By fine-tuning their contributions, we can harness the strengths of each LLM for specific subtasks within a larger task. Second, it allows the model to adapt to changes in the data or the task itself. As new data is introduced or the problem evolves, individual LLMs’ contributions can be adjusted accordingly, ensuring that the ensemble remains effective and up-to-date. However, these ensembles ignore the following key elements:

\begin{itemize}
    \item \textit{External Knowledge Integration:} The approach involves integrating external knowledge sources, such as Knowledge Graphs (KGs) and Clinical Practice Guidelines, into the LLM ensemble. These sources provide additional context and information that can enhance the quality of the generated text.
    \item \textit{Reward Functions:} The external knowledge is not simply added as static information but is used as reward functions during the ensembling process. In simpler terms, this means the ensemble of models gets rewarded when they produce text that matches or incorporates external knowledge. This reward system promotes logical consistency and meaningful connections with that knowledge.
    \begin{itemize}
        \item \textit{Logical Coherence:} By incorporating external knowledge, the ensemble of LLMs aims to produce a more logically coherent text. It ensures the generated content aligns with established facts and relationships in the external knowledge sources.
        \item \textit{Semantic Relatedness:} The ensemble also focuses on improving the semantic relatedness of the generated text. This means that the text produced by the LLMs is factually accurate, contextually relevant, and meaningful.
    \end{itemize}
\end{itemize}
Such attributes are important when LLMs are designed for critical applications like Motivational Interviewing \cite{sarkar2023review}. Motivational interviewing is a communication style often used in mental health counseling, and ensuring logical coherence and semantic relatedness in generated responses is crucial for effective interactions \cite{shah2022modeling}. 
\subsubsection{Deep Ensemble of LLMs} introduces an innovative approach using NeSy-AI, in which e-LLMs are fine-tuned with the assistance of an evaluator. This evaluator comprises constraints and graph-based knowledge representations and offers rewards to guide the generation of e-LLMs based on the aforementioned properties. Concurrently, it incorporates knowledge source concepts in the form of representations to compel e-LLMs to include and prioritize these concepts, enhancing their reliability (refer to Figure \ref{fig:7} for illustration). Another key objective of the deep ensemble approach is to transform e-LLMs into a Mixture of Experts \cite{artetxe2022efficient} by enhancing individual LLMs through a performance maximization function \cite{kwon2022reward}.

\subsection{Explainability and User-level Explainable LLMs (UExMs)}
Achieving effective and human-understandable explanations from LLMs or even from their precursor language models (LMs) remains complex. Previous attempts to elucidate BlackBox LMs have utilized techniques like surrogate models (such as LIME \cite{ribeiro2016should}), visualization methods, and adversarial perturbations to the input data \cite{chapman2021fimap}. While these approaches provide explanations, they operate at a relatively basic level of detail, which we have referred to as system-level explainability \cite{gaur2022knowledge}.

System-level Explainability has been developed under the purview of post-hoc Explainability techniques that aim to interpret the attention mechanism of LMs/LLMs without affecting their learning process. These techniques establish connections between the LM's attention patterns and concepts sourced from understandable knowledge repositories. Within this approach, two methods have emerged: (a) Attribution scores and LM Tuning \cite{slack2023explaining}  and Factual Knowledge-based Scoring and LM Tuning \cite{yang2023chatgpt, sun2023thinkongraph}. The latter method holds particular significance in the domain of health and well-being because it focuses on providing explainability for clinicians as users. This method relies on KGs or knowledge bases like the Unified Medical Language System (UMLS) \cite{bodenreider2004unified}, SNOMED-CT \cite{PMID:17095826}, or RXNorm \cite{10.1136/amiajnl-2011-000116} to enhance its functionality.

While the post-hoc method can provide explanations (by modeling it as a dialogue system \cite{lakkaraju2022rethinking}), it does not guarantee that the model consistently prioritizes essential elements during training \cite{jiang2021can}. Its explanations may be coincidental and not reflect the model's actual decision-making process. More recently, the focus has shifted to ``explainability by design,'' particularly in critical applications like healthcare. A recent example is the Transparency and Interpretability Framework for Understandability (TIFU), proposed by \citet{joyce2023explainable}, which connects inherent explainability to a higher level of explainability in the mental health domain.  The primary motivation for pursuing such an explainability, called User-level explainability, is to ensure that healthcare professionals and patients are given contextually relevant explanations that help them understand the AI system’s process and outcomes so they can develop confidence in AI tools.

\begin{tcolorbox}[colback=blue!5!white,colframe=blue!75!black]
A User-level Explainability in LLMs implies that humans can rely on the AI system to the extent that they can reduce the need for human oversight, monitoring, and verification of the system's outputs. To trust a deployed LLM, we must have adequate insight into how it generates an output based on a given input.
\end{tcolorbox}

\begin{tcolorbox}[colback=blue!5!white,colframe=blue!75!black, title=UExMs]
UExMs provide user-explainable insights by utilizing expert-defined instructions, statistical knowledge (attention), and knowledge retriever.
\end{tcolorbox}

UExMs can be practically realized in four different ways:
\subsubsection{UExMs with Generating Evaluator Pairing:} This defines a generative and evaluator-based training of UExMs where any LLM is paired with a knowledge-powered evaluator, either accelerates or deaccelerates the training of LLMs, depending on whether the final generation is within the acceptable standards of the evaluator. ``\textit{On the weekend, when I want to relax, I am bothered by trouble concentrating while reading the newspaper or watching television. Need some advice}'' clearly indicates that the individual is experiencing specific issues related to concentration during leisure time. This query is more than just a casual comment; it highlights a problem that is affecting the user's ability to unwind effectively. Now, consider the two scenarios: 
\begin{itemize}
    \item \textit{Without an Evaluator (Generic Response):} In the absence of an evaluator, an LLM might provide a generic set of activities or advice, such as “practice mindfulness, limit distractions,  break tasks into smaller chunks,” and so on. While this advice is generally useful for improving concentration, it lacks the depth and specificity needed to address the user's potential underlying issues.
    \item \textit{With an Evaluator (Specific Response):} When integrated into the LLM, an evaluator can analyze the user's query more comprehensively. In this case, the evaluator can recognize that the user's difficulty concentrating during relaxation may indicate an underlying sleep-related issue. Considering this possibility, the language model can provide more targeted and informed advice.
    
    For instance, the evaluator might suggest asking further questions like: (a) Do you have trouble sleeping at night? (b) How much sleep do you typically get on weekends? (c) Have you noticed other sleep-related symptoms, such as daytime drowsiness? (d) Have you considered the possibility of a sleep disorder? By incorporating an evaluator, the LLM can guide the conversation toward a more accurate understanding of the user's situation. To put it simply, the LLM, when assisted by an evaluator, will provide a coherent answer that encompasses all aspects of the user's question \cite{gaur2022iseeq, gaur2023dynamic}. Further, the evaluator prevents the model from generating hallucinated, off-topic, or overly generic responses. A framework like ISEEQ integrates generator and evaluator LLMs for generating tailored responses in general-purpose and mental health domains \cite{gaur2022iseeq}. Additionally,  PURR and RARR contribute to refining segments of LLM design aimed at mitigating hallucination-related problems in these models \cite{chen2023purr, gao2023rarr}.
\end{itemize}

\begin{figure*}[t]
    \centering
    \includegraphics[width=\textwidth]{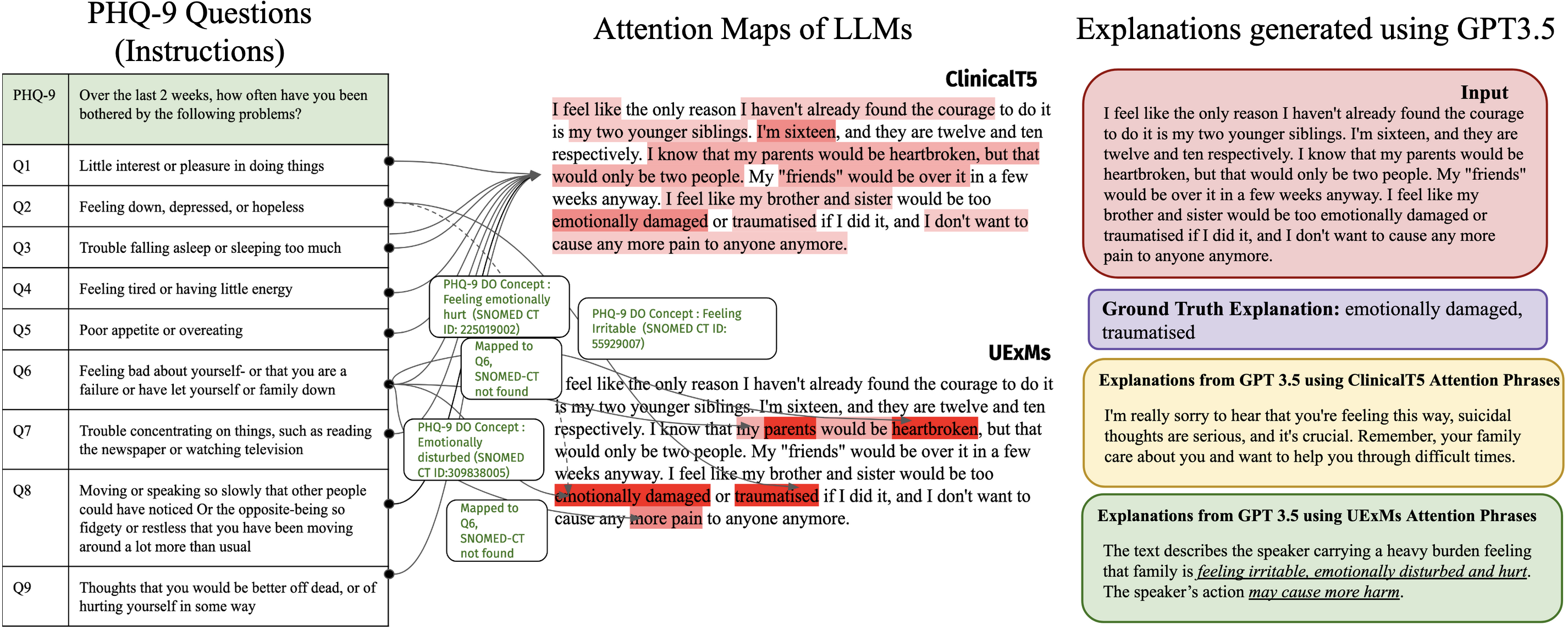}
    \caption{An instance of user-level explainability in a UExM is when the model uses questions from PHQ-9 to guide its actions and relies on SNOMED-CT, a clinical knowledge base, to simplify complex concepts (concept abstraction). This approach helps the model offer explanations that closely align with the ground truth. PHQ9-DO: PHQ-9-based Depression Ontology.}
    \label{fig:4}
\end{figure*}

\noindent To illustrate this concept, refer to Figure \ref{fig:4}, which illustrates a task where a generative LM takes user input and provides an assessment in natural language, specifically within the PHQ-9 context \cite{dalal2023cross}. The figure shows two LLMs: ClinicalT5-large, a powerful LM with 38 billion parameters, and UExM, which is essentially ClinicalT5-large but enhanced with a PHQ-9-grounded evaluator. This demonstrates that by employing an evaluator with predefined questions, we can assess how well the attention of generative ClinicalT5-large aligns with those specific questions. This approach helps ensure that the generated explanations are relevant and comprehensive, making them clinically applicable, particularly when healthcare professionals rely on standardized guidelines like the PHQ-9 to evaluate patients for depression \cite{honovich2022true}.

\subsubsection{UExMs with Retriever Augmentation and Process Knowledge:} It's commonly observed that the process of generating responses by LLMs lacks transparency, making it difficult to pinpoint the origin of their answers. This opacity raises questions about how the model derives its responses.
\begin{itemize}
    \item \textit{The emergence of Retrieval-Augmented Generation LMs:} A novel class of LMs has surfaced to tackle this issue and add a layer of supervision to language model outputs. Examples include REALM \cite{guu2020retrieval}, LAMA \cite{petroni2019language}, ISEEQ \cite{gaur2022iseeq}, and RAG \cite{lewis2020retrieval}, which integrate a generator with a dense passage retriever and access to indexed data sources. LLMs with retrieval-augmented architectures have started to show understandable and accountable responses \cite{lyu2023improving}. For instance,  GopherCite \cite{menick2022teaching} and NeMo Guardrails \cite{rebedea2023nemo} are LLMs that leverage a knowledge base to supply supporting evidence for nearly every response generated by the underlying LLM. 
    \item \textit{The emergence of Process Knowledge-guided Generation LMs:} Process Knowledge refers to guidelines or instructions created by experts in a domain \cite{roy2023process}. For instance, in mental health, PHQ-9 is the process of knowledge for screening depression \cite{kroenke2001phq}, NIDA’s Attention Deficiency Hyperactivity Disorder Test, and the World Health Organization’s Wellness Indices \cite{topp20155}. The questions in these guidelines can act as rewards for enriching latent generations (e.g., answerability test \cite{yao2023development}) \cite{hagendorff2023machine}. 
\end{itemize}

\subsubsection{UExMs with Abstention} While a retriever has been integrated into an LLM, it doesn't guarantee meaningful explainability. When considering a ranked list of retrieved and expanded documents, an LLM is still vulnerable to generating incorrect or irrelevant explanations. Therefore, it's crucial to eliminate meaningless hidden generations before they are converted into natural language. For example, the ReACT framework employs Wikipedia to address spurious generation and explanations in LLMs \cite{yao2022react}. However, it relies on a prompting method rather than a well-grounded domain-specific approach, which can influence the generation process used by the LLM \cite{yang2023large}. Alternatively, pruning methods and an abstention rule have also been used to reduce irrelevant output from LLMs. A more robust approach would involve utilizing procedural or external knowledge as an evaluator guiding LLM-generated content that enhances meaningful understanding.

\subsection{Safety}
\begin{tcolorbox}[colback=blue!5!white,colframe=blue!75!black]
Safety and explainability are closely intertwined concepts for AI systems. While a safe AI system will inherently demonstrate explainability, the reverse isn't necessarily true; an explainable system may or may not be safe.
\end{tcolorbox}

Recently, there has been a proliferation in safety-enabled research, particularly in LMs and LLMs. \citet{perez2022red} performed red-teaming between LMs to determine if an LM can produce harmful text. The process did not include humans in generating these adversarial test cases. Further, the research did not promise to address all the critical safety oversights comprehensively; instead, it aimed to spotlight instances where LMs might exhibit unsafe behavior. \citet{scherrer2023evaluating} delves more deeply into the safety issues in LLMs by examining their behavior in moral scenarios. The study found that LLMs only focus on generating fluent sentences and overlook important words/concepts contributing to stable decisions. Further, datasets like DiSafety and SafeTexT are designed to induce safety in LMs/LLMs through supervised learning \cite{meade2023using, levy2022safetext}. These discussions surrounding safety gained heightened attention, particularly within the National Science Foundation (NSF), leading to the launch of two programs: (a) Safety-enabled Learning and (b) Strengthening AI. In a recent webinar, NSF outlined three fundamental attributes of ensuring safety: grounding, instructability, and alignment\footnote{https://new.nsf.gov/funding/opportunities/national-artificial-intelligence-research}.

\subsubsection{Grounding:} 
\begin{figure}
    \centering
    \includegraphics[width=\linewidth]{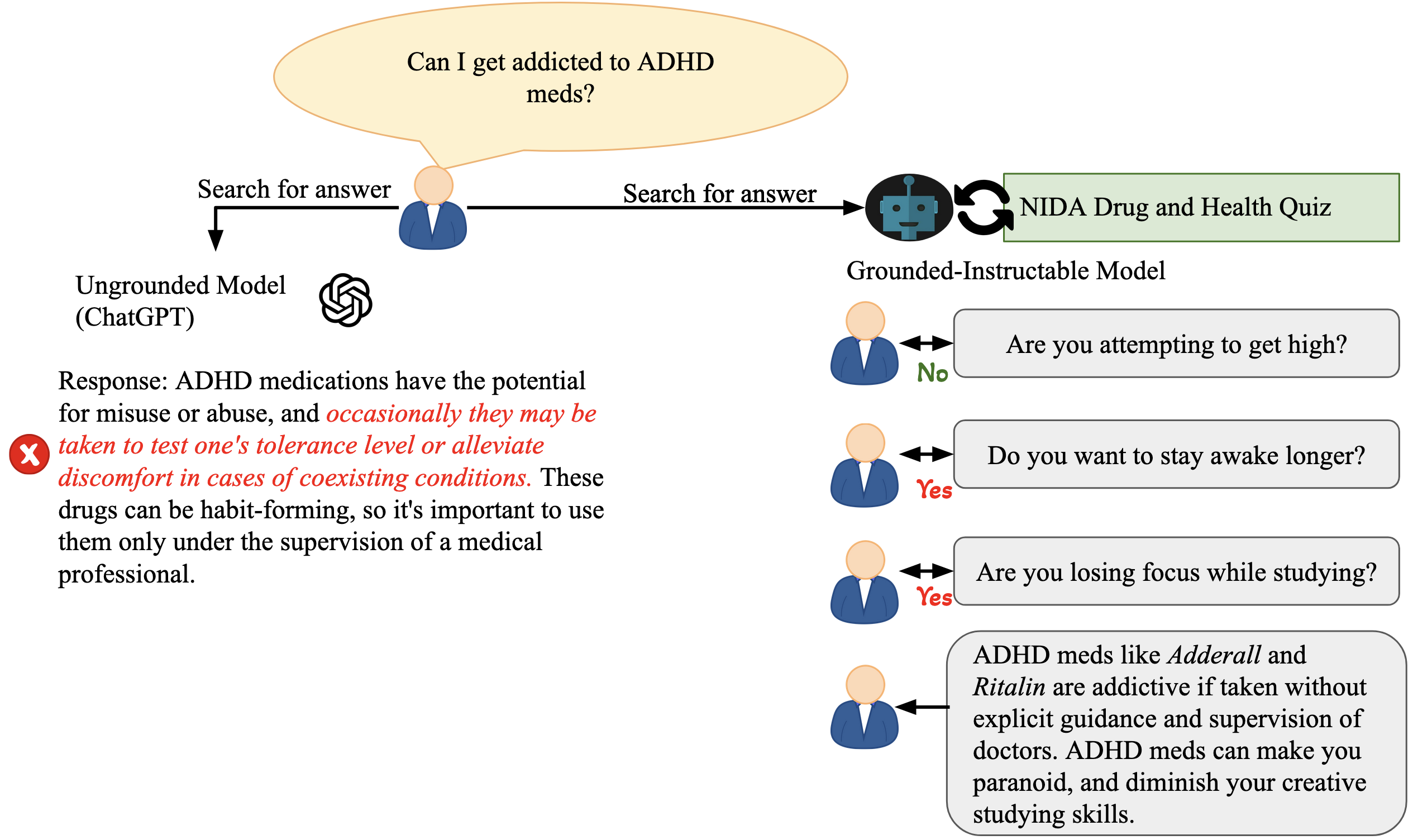}
    \caption{An Illustration of grounding and instruction-following behavior in an LLM (right) tuned with support from health and well-being-specific guidelines. ChatGPT’s response was correct, but it isn’t safe.}
    \label{fig:5}
\end{figure}

In essence, groundedness is the foundation upon which both explainability and safety rest. Without a strong grounding in the provided instructions, the AI may produce results that stray from the desired outcome, potentially causing unintended consequences. For instance, consider the scenario depicted in Figure \ref{fig:5}. An LLM that isn’t grounded in domain-specific instruction, like the ChatGPT, results in an unsafe response. On the other hand, a relatively simple LLM, like T5-XL, tuned by grounding in domain-specific instructions, attempts to ask follow-up questions to gather the necessary context for a coherent response. The changes in T5-XL's behavior due to the NIDA\footnote{National Institute on Drug Abuse} quiz highlight the importance of being able to instruct and align AI, which is key for safety\footnote{https://psychcentral.com/quizzes/adhd-quiz}. 

\subsubsection{Instructability:} In the context of AI safety, instructability encompasses the assurance that the AI understands and complies with user preferences, policies, and moral beliefs. Making the LMs bigger and strengthening the rewards makes the models power-hungry rather than ethical and safe. For instance, the guardrails instantiated for the safe functioning in OpenAI’s ChatGPT, the rules within DeepMind’s Sparrow, and the list of rules within Anthropic’s Claude cannot reliably prove that they are safe.

The idea of having systems that follow instructions has been around since 1991, mainly in robotics and, to some extent, in text-based agents. It's crucial because it helps agents learn tasks, do them well, and explain how they did it, making sharing knowledge easier between humans and AI and showing they can follow human instructions. One way to do this is by using grounded instruction rules, especially in the field of mental health. Clinical practice guidelines like PHQ-9 for depression and GAD-7 for anxiety, with their questions, can serve as instructions for AI models focused on mental health. Grounded rules have two key benefits for safety. First, they tend to be helpful and harmless, addressing a common challenge for AI models. Second, they promote absolute learning, avoiding tricky trade-off situations.

\subsubsection{Alignment:} When we talk about alignment in LMs, it means ensuring that even a model designed to follow instructions doesn't produce unsafe results \cite{macdonald1991instructable}. This can be a tricky problem, as discussed in Nick Bostrom's book ``\textit{Superintelligence},'' where it's called ``\textit{perverse instantiations}'' \cite{10.5555/2678074}. This happens when the LM/LLMs figure out how to meet a goal, but it goes against what the user wants \cite{ngo2022alignment}. So, the challenge is to create an AI that follows instructions and finds the best way to achieve a goal while keeping users happy, a concept referred to as ``\textit{Wireheading}'' in ``\textit{Superintelligence}.'' Following are perspectives on why it happens and what can be done: 

\begin{itemize}
    \item \textit{Context Awareness (CA) and Contextual Rewards (CR):} CA refers to the training of LMs/LLMs to focus on words or phrases that have direct translation to concepts in factual knowledge sources. CR serves the function of facilitating CA. They achieve this by incorporating evaluator modules that analyze the hidden or latent representations within the model with respect to the concepts present in the knowledge sources. CR reinforces and guides CA by rewarding the model when it correctly identifies and incorporates knowledge-based concepts into its responses.
    \item \textit{Misalignment in latent representations caused by misleading reward associations:} We acknowledge the inherent perceptiveness of LMs and LLMs, a quality closely linked to the quantity of training data they are exposed to. Nevertheless, having a larger training dataset leads to superior performance scores, but it may not necessarily meet the expectations of human users. Bowman has demonstrated that a model achieving an F1 score of over 80\% still struggles to prioritize and pay adequate attention to the concepts users highly value \cite{bowman2023eight}. This happens because optimization algorithms and attention methods in LLMs can attempt to induce fake behavior. Further, if the rewards specified are not unique to the task but rather general, the model will have difficulty aligning with desired behaviors \cite{shah2022goal}.
    \item \textit{Deceptive Alignment during Training:} Spurious reward collections can lead to deceptive training. It is important to train the LMs/LLMs with paraphrases and adversarial input while examining the range of reward scores and the variations in the loss functions. If LMs/LLMs demonstrate high fluctuations in the rewards and the associated effect on loss, it would most likely result in brittleness during deployment. Methods like the chain of thoughts and the tree of thoughts prompting can act as sanity checks to examine the deceptive nature of LMs/LLMs \cite{alignmentforumCognitiveEmulation, yao2023tree}.
\end{itemize}

\begin{tcolorbox}[colback=blue!5!white,colframe=blue!75!black, title=Brief Summary]
Knowledge of the AI system and domain is pervasive in achieving consistency, reliability, explainability, and safety for building a \textbf{T}rustworthy AI system. 
\begin{itemize}
    \item For \textbf{C}onsistency, rules, and knowledge can make LLMs understand and fulfill user expectations confidently
    \item \textbf{R}eliability is ensured by utilizing the rich knowledge contained in KGs to empower an ensemble of LLMs to produce consistent and mutually agreeable results with high confidence.
    \item For \textbf{E}xplainability, LLMs use their knowledge, retrieved knowledge, and rules that were followed to attain consistency and reliability to explain the generation effectively. 
    \item \textbf{S}afety in LLMs is upheld by consistently grounding their generation and explanations in domain knowledge and assuring the system's adherence to expert-defined rules or guidelines.
\end{itemize}
\end{tcolorbox}

\section{The CREST Framework}
To realize CREST, we now provide succinct descriptions of its key components and highlight open challenges for AI and NeSy-AI communities in NLP (see Figure \ref{fig:6}). We delve into three components of the CREST framework in the following subsections:

\begin{figure*}[t]
    \centering
    \includegraphics[width=\textwidth]{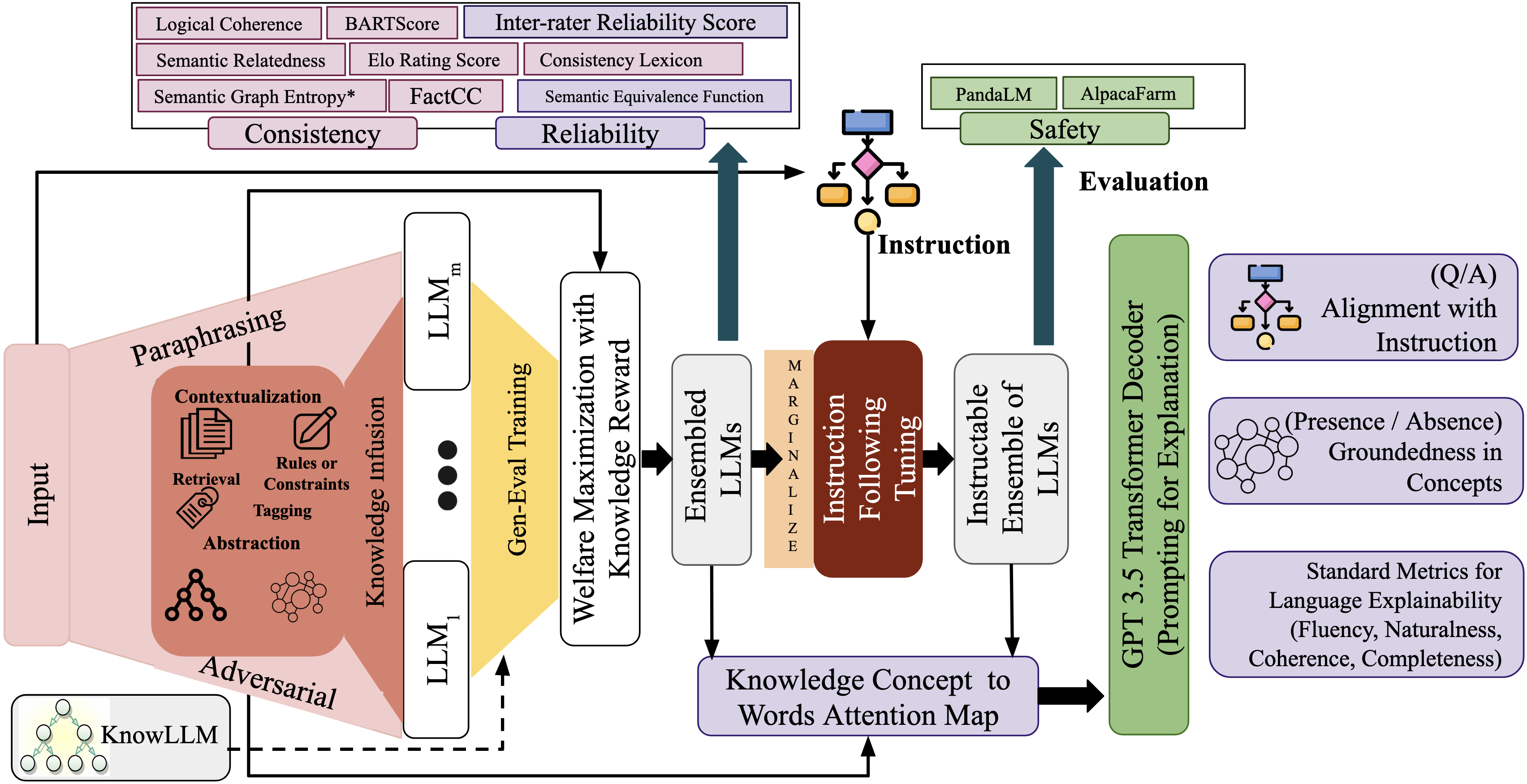}
    \caption{The CREST framework operationalizes ``explainability and safety'' by ensuring the model is reliable and consistent. LLMs (1 to m) can be replaced with LLMs in Figure \ref{fig:2}, and the knowledge used in infusion refers to UMLS and SNOMED-CT for a clinical domain, as we examined CREST for mental health. Gen-Eval: Generator and Evaluator pairing. KnowLLM: LLMs created using KGs.}
    \label{fig:6}
\end{figure*}

\subsection{NeSy-AI for Paraphrased and Adversarial Perturbations}
Paraphrasing serves as a technique to enhance an AI agent's calibration by making it aware of the different ways an input could be expressed by a user \cite{du2023incorporating}. This, in turn, contributes to increasing the AI agent's consistency and reliability. Agarwal et al. introduced a pioneering NeSy AI-based approach to paraphrasing. In their method, they employed CommonSense, WordNet, and Wikipedia knowledge graphs to generate paraphrases that held equivalent meanings but were perceived as distinct by the AI agent \cite{agarwal2023towards}. However, there are some promising directions for NeSy paraphrasing. First is contextualization, which involves augmenting the input with meta-information retrieved from a rank list of documents. This transforms NLP's not-so-old question rewriting problem into a knowledge-guided paraphrasing method. The second is abstraction, which involves identifying the function words (e.g., noun phrases, verb phrases) and named entities and replacing them with abstract concepts. For instance, the following sentence, ``\textit{Why trauma of harassment is high in $boys|girls$?}'' is abstracted to ``\textit{why trauma of (harassment $\rightarrow$ mistreatment) is high in ($boys|girls$ $\rightarrow$ students)?}''.  Both of these methods can benefit from existing learning strategies of LLMs, such as marginalization \cite{wang2022self} and reward-based learning \cite{jie2023prompt}.

NeSy-AI for adversarial perturbations (AP)  uses general-purpose KGs to carefully change the sentence to examine the brittleness in LLMs’ outcomes.
\begin{tcolorbox}[colback=red!5!white,colframe=red!75!black, title=Example of Adversarial Generation using NeSy-AI]
S1: I have been \textbf{terrible} in battling with my loneliness. My overly introvertedness and \textbf{terrible} choice of few friends are the reasons for who I am. The only part I considered funny in this situation was that none of my friends knew how I felt. It seems they are \textbf{childish}.
\tcblower
S1-AP: I have been \ul{\textbf{horrible}} at battling my loneliness. My overly introvertedness and \ul{\textbf{horrible}} choice of few friends are the reasons for who I am. The only part I regarded as \ul{\textbf{sarcastic}} in this situation was that none of my friends knew how I felt. It seems they are \ul{\textbf{youngsters}}.
\end{tcolorbox}
The  Flan T5 (11B) estimates S1 to have a ``\textit{negative}'' sentiment with a confidence score of 86.6\% and S1-AP  to have a ``\textit{positive}'' sentiment with a 61.8\% confidence score.  The confidence scores are predicted probability estimates. LLMs must concentrate on the contextual notions (such as loneliness and introversion) and the abstract meaning that underlies both S1 and S1-AP—that is, the influence on mental health and well-being—to attain consistency and reliability in such inadvertent settings.

\subsection{Knowledge-infused Ensembling of LLMs}

As mentioned above, e-LLMs have many benefits; however, simply statistical methods of ensembling, which consist of averaging the outcomes from black box LLMs, do not make an ensembled LLM consistent and reliable. Knowledge-infused Ensemble represents a particular methodology where the knowledge (general purpose or domain-specific) modulates the latent representations of the LLMs to yield the best of world outcomes. This can happen in one of three ways:

\begin{enumerate}
    \item \textit{LLMs over KGs (KnowLLMs):} Similar to the process of training any LLM on text documents, which involves formulating it as a task of predicting the next word in a sentence, KnowLLMs undertake the training of LLMs using a variety of KGs such as CommonSense, Wikipedia, and UMLS. In KnowLLMs, the training objective is redefined as an autoregressive function over $<subject><predicate><object>$ coupled with pruning based on existing state-of-the-art KG embedding methods. Introducing pruning is crucial in KnowLLMs to prevent the model from making unwarranted inferences and forming incorrect links. This is vital for ensuring the safety and trustworthiness of the knowledge generated by KnowLLMs. In other words, by pruning, KnowLLMs can filter out irrelevant or potentially misleading information, thereby enhancing the quality of their responses and minimizing the risk of spreading false or harmful knowledge.
    \item \textit{Generative Evaluator Tuning:} This approach suggests using reinforcement learning to improve the training of e-LLMs. It combines the traditional training method with rewards from KnowLLMs, which act as extra guidelines. These rewards encourage the e-LLM to generate text that aligns with specific desired characteristics, such as mental health concepts. If the e-LLM's output doesn't meet these criteria or is logically incorrect according to KnowLLM, it receives negative rewards, even if it's similar to the ground truth based on similarity scores. This method helps e-LLMs produce more contextually relevant and accurate text.
    \item \textit{Instruction Following Tuning:}  Instruction Tuning has recently emerged as a promising direction to teach LLMs to match the expectations of humans. Though promising, it requires a substantial amount of samples, and there is no perfect quantifiable method to measure the ``\textit{instruction following}'' nature of LLMs. And, if we decide to embark on a ``\textit{mixture of experts}'' like e-LLMs, it would be hard to make separate procedures for instruction tuning over e-LLMs. Thus, we take inspiration from Process Knowledge-infused Learning, a mechanism for intrinsically tuning the LMs or an ensemble of LMs.  Roy et al. demonstrated how questionnaires in the clinical domain, which can be considered a constraint, can enable LMs to generate safe and consistently relevant questions and responses \cite{roy2023process}. This approach works on a simple Gumble Max function, which allows structural guidelines to be used in the end-to-end training of LMs. This approach is fairly flexible for ``\textit{instruction-following-tuning}'' of e-LLMs and ensuring the instruction is followed. 
\end{enumerate}

\subsection{Assessment of CREST}
The CREST framework significantly emphasizes incorporating knowledge and utilizing knowledge-driven rewards to support e-LLMs in achieving trust. To assess the quality of e-LLMs' output, it's crucial to employ metrics that account for the knowledge aspect. For instance, the logical coherence metric evaluates how well the content generated by e-LLMs aligns with the flow of concepts in KGs and context-rich conversations. Additional metrics like Elo Rating \cite{zheng2023judging}, BARTScore \cite{liu2023gpteval}, FactCC \cite{kryscinski2020evaluating}, and Consistency lexicons can be improved to account for the influence of knowledge on e-LLMs' generation. However, when it comes to assessing reliability, aside from the established Cohen's or Fleiss Kappa metrics, an effective alternate metric is not available.

Safety aspects in CREST are best evaluated when knowledge-tailored e-LLMs are instructed to adhere to guidelines established by domain experts. Existing metrics like PandaLM \cite{wang2023pandalm} and AlpacaFarm \cite{dubois2023alpacafarm} are based on LLMs, which themselves may exhibit vulnerabilities to unsafe behaviors. While such metrics may be suitable for open-domain applications, when it comes to critical applications, safety metrics must be rooted in domain expertise and align with the expectations of domain experts.

In CREST, explainability is evaluated through two approaches requiring expert verification and validation. One method involves analyzing the ``\textit{Knowledge Concept to Word Attention Map}'' to gain insights into CREST's reasoning process and verify whether the model's decisions align with domain knowledge and expectations \cite{gaur2018let}. Another method involves using knowledge concepts and domain-specific decision guidelines (e.g., clinical practice guidelines) to enable LLMs like GPT 3.5 to generate human-understandable explanations (as shown in Figure \ref{fig:4}).

\subsection{A Case Study in Mental Health in Brief}
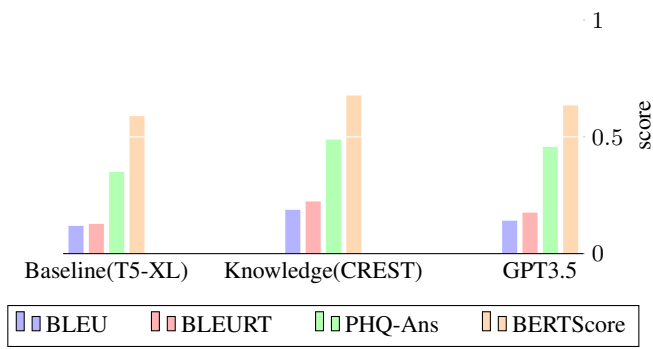
\begin{figure}
\small
\begin{tikzpicture}
  \centering
  \begin{axis}[
        ybar, axis on top,
        height=5cm, width=8.5cm,
        bar width=0.2cm,
        ymajorgrids, tick align=inside,
        major grid style={draw=white},
        enlarge y limits={value=.1,upper},
        ymin=0, ymax=1,
        axis x line*=bottom,
        axis y line*=right,
        y axis line style={opacity=0},
        tickwidth=0pt,
        enlarge x limits=true,
        legend style={
            at={(0.5,-0.2)},
            anchor=north,
            legend columns=-1,
            /tikz/every even column/.append style={column sep=0.5cm}
        },
        ylabel={score},
        symbolic x coords={
           Baseline(T5-XL), Knowledge(CREST), GPT3.5},
       xtick=data
    ]
    \addplot [draw=none, fill=blue!30] coordinates {
      (Baseline(T5-XL),0.1188)
      (Knowledge(CREST), 0.1877) 
      (GPT3.5, 0.1412)};
   \addplot [draw=none,fill=red!30] coordinates {
      (Baseline(T5-XL),0.1274)
      (Knowledge(CREST), 0.2234) 
      (GPT3.5, 0.1756)};
   \addplot [draw=none, fill=green!30] coordinates {
      (Baseline(T5-XL),0.35)
      (Knowledge(CREST), 0.4886) 
      (GPT3.5,0.4571)};
    \addplot [draw=none, fill=orange!30] coordinates {
      (Baseline(T5-XL), 0.5889)
      (Knowledge(CREST), 0.6772) 
      (GPT3.5,0.6344)};
      
    \legend{BLEU, BLEURT, PHQ-Ans, BERTScore}
  \end{axis}
  \end{tikzpicture}
    \caption{The CREST findings on the PRIMATE dataset include PHQ-9 answerability, calculated as the mean Matthew Correlation Coefficient score. This score is computed by comparing predicted Yes/No labels against the ground truth across nine PHQ-9 questions. BLEURT score is computed between questions generated by LLMs and PHQ-9 questions \cite{sellam2020bleurt}. LLMs were prompted to create questions based on sentences identified as potential answers to the PHQ-9 questions. PHQ-Ans: PHQ-9 Answerability.}
    \label{fig:7}
\end{figure}

We present a preliminary performance of CREST on the PRIMATE dataset, introduced during ACL's longstanding Clinical Psychology workshop \cite{gupta2022learning}. It is a distinctive dataset designed to assess the LM's ability to consistently estimate an individual's level of depression and provide yes/no responses to PHQ-9 questions, which is a measure of its reliability. Figure \ref{fig:7} shows the performance of CREST and knowledge-powered CREST relative to GPT 3.5. Including knowledge in CREST showed an improvement of 6\% in PHQ-9 answerability and 21\% in BLEURT over GPT 3.5, which was used through the prompting method. The e-LLMs in CREST were Flan T5-XL (11B) and T5-XL (11B).

\section{Conclusion and Future Work}
LLMs and broadly generative AI represent the most exciting current approach but are not the solution for Trustworthy AI alone.  LLMs exhibit undesired behaviors during tasks such as question answering, making them susceptible to threats and resultant problematic actions. Therefore, there is a need for innovative approaches to identify and mitigate threats posed both to LLMs and by LLMs to humans, especially when they are to be used for critical applications such as those in health and well-being.  A comprehensive solution is needed beyond the implementation of guardrails or instruction adjustments. This solution should encourage LLMs to think ahead, leveraging domain knowledge for guidance. The CREST framework offers a promising approach to training LLMs with domain knowledge, enabling them to engage in anticipatory thinking through techniques like paraphrasing, adversarial inputs, knowledge integration, and fine-tuning based on instructions.

We presented a preliminary effort in implementing the CREST framework that yields enhancements over GPT3.5 on PRIMATE, a PHQ-9-based depression detection dataset. We plan to experiment with CREST on knowledge-intensive language generation benchmarks, like HELM \cite{liang2022holistic}. Further, we plan on automating user-level explanations without dependence on pre-trained LLMs (e.g., GPT3.5). Our future endeavors involve developing more effective training methodologies for e-LLMs powered by the CREST framework. Additionally, we will incorporate robust paraphrasing and adversarial generation techniques to assess the consistency and reliability of e-LLMs when they are exposed to knowledge. This will also open avenues for further research into crafting quantitative metrics that evaluate reliability, safety, and user-level explainability. 

\section{Acknowledgement}
We express our gratitude to Drs. Amitava Das and Valerie L. Shalin for their invaluable reviews and insightful suggestions on the manuscript. We acknowledge partial support from the NSF EAGER award \#2335967 and the UMBC Summer Faculty Fellowship. Any opinions, conclusions, or recommendations expressed in this material are those of the authors and do not necessarily reflect the views of the NSF or UMBC. 
% Thanks to Valerie Shalin, Amitava Das
% UMBC Summer Faculty Fellowship

%\bibliographystyle{aaai22}
\bibliography{aaai22.bib}

\end{document}